\documentclass{article} % For LaTeX2e
\usepackage{nips13submit_e,times}
\usepackage{url}

\usepackage[algo2e,ruled]{algorithm2e}
\usepackage{graphicx} % more modern
\usepackage{hyperref,multirow,float}
\usepackage{paralist,xspace}
\usepackage{subfigure}
\usepackage{steinmetz}
\usepackage{amsmath,amssymb}

\title{A Parallel SGD method with Strong Convergence}

\author{
Dhruv Mahajan \\
Microsoft Research \\
Bangalore, India \\
\texttt{dhrumaha@microsoft.com} \\
\And
S. Sathiya Keerthi\\
Microsoft Corporation \\
Mountain View, USA \\
\texttt{keerthi@microsoft.com}
\AND
S. Sundararajan \\
Microsoft Research \\
Bangalore, India \\
\texttt{ssrajan@microsoft.com} \\
\And
L\'{e}on Bottou\\
Microsoft Research \\
New York, USA \\
\texttt{leonbo@microsoft.com}\\
}

\nipsfinalcopy

\begin{document}

\maketitle

\begin{abstract}
This paper proposes a novel parallel stochastic gradient descent (SGD) method that is obtained by applying parallel sets of SGD iterations (each set operating on one node using the data residing in it) for finding the direction in each iteration of a batch descent method. The method has strong convergence properties. Experiments on datasets with high dimensional feature spaces show the value of this method.
\end{abstract}

%\maketitle

\def\grad{\nabla}
\def\wtilde{\tilde{w}}
\def\Cone{{\cal{C}}^1}
\def\kappap{\kappa^\prime}
\def\Lhat{\hat{L}}
\def\fhat{\hat{f}}
\def\what{\hat{w}}
\def\dhat{\hat{d}}
\def\mysgn{\operatorname{sgn}}
\def\ftilde{\tilde{f}}
\def\khat{\hat{k}}
\def\defs{\stackrel{\text{def}}{=}}
\def\vhat{\hat{v}}

\def\ttilde{\tilde{t}}
\def\that{\hat{t}}
\def\tstar{t^\star}

{\bf Introduction.}
We are interested in the large scale learning of linear classifiers. Let $\{x_i,y_i\}$ be the training set associated with a binary classification problem ($y_i\in\{1,-1\}$). Consider a linear classification model, $y=\mysgn(w^Tx)$. Let $l(w\cdot x_i,y_i)$ be a continuously differentiable, non-negative, convex loss function that has Lipschitz continuous gradient. This allows us to consider loss functions such as least squares, logistic loss and squared hinge loss. Hinge loss is not covered by our theory since it is non-differentiable. Our aim is to to minimize the regularized risk functional $f(w)=\frac{\lambda}{2}\|w\|^2 + L(w)$ where $\lambda>0$ is the regularization constant and $L(w)=\sum_i l(w\cdot x_i,y_i)$ is the total loss. The gradient function, $g=\grad f$ is Lipschitz continuous.

For large scale learning on a single machine, it is now well established that example-wise methods\footnote{These methods update $w$ after scanning each example.} such as stochastic gradient descent (SGD) and its variations~\cite{bottou2010, leroux2012, johnson2013} and dual coordinate ascent~\cite{hsieh2008} are much faster than batch gradient-based methods for reaching weights with sufficient training optimality needed for attaining steady state generalization performance. However, example-wise methods are inherently sequential.

For tackling problems involving huge sized data, distributed solution becomes necessary. One approach to parallel SGD solution~\cite{zinkevich2010} is via (iterative) parameter mixing~\cite{mann2009,hall2010}. Consider a distributed setting with a master-slave architecture\footnote{An {\it AllReduce} arrangement of nodes~\cite{agarwal2011} may also be used.} in which the examples are partitioned over $P$ slave computing nodes. Let: $I_p$ be the set of indices $i$ such that $(x_i,y_i)$ sits in the $p$-th node; and $L_p(w) = \sum_{i\in I_p} l(w\cdot x_i,y_i)$ be the total loss associated with node $p$. Thus, $f(w)=\frac{\lambda}{2} \|w\|^2 + \sum_p L_p(w)$. Suppose the master node has the current weight vector $w^r$ and it communicates it to all the nodes. Each node $p$ can form the approximation,
\begin{equation}
\ftilde_p=\frac{\lambda}{2} \|w\|^2 + L_p(w)
\label{ftilde}
\end{equation}
of $f$ using only its examples, and do several SGD epochs (local passes over its examples) on $\ftilde_p$ and reach a point $w_p$. The $w_p\;\forall p$ can then be communicated back to the master node and averaged to form the next iterate $w^{r+1}$. One can stop after just one major iteration (go from $r=0$ to $r=1$) or repeat many such major iterations. Convergence theory for such methods is limited, and, even that requires a complicated analysis~\cite{zinkevich2010}. There are two main issues related to variance and bias: (a) When the number of nodes is large, the $\ftilde_p$ are very different from each other, and so, the variability in the $w_p$ is large and the averaged weight vector is also far away from $w^\star=\arg\min f(w)$. (b) If we use too many SGD epochs within each node $p$, then, within each major iteration, SGD will converge to the minimizer of $\ftilde_p$ irrespective of the starting point $w^r$, making the major iterations useless.

{\bf A New Parallel SGD method.}
Our main idea is to use a descent method for batch training and, in each of its iterations, compute the direction by doing SGD iterations in parallel on function approximations that are ``better" than the $\ftilde_p$. We begin each iteration $r$ by computing the gradient $g^r$ at the current point $w^r$.\footnote{It is worth noting that recently proposed powerful SGD methods~\cite{johnson2013} also compute full batch gradients once every few SGD epochs.} One can communicate $w^r$ and $g^r$ to all $P$ (slave) nodes. The direction $d^r$ is formed as follows. Each node $p$ constructs an approximation of $f(w)$ using only information that is available in that node\footnote{This information includes the examples sitting in the node as well as $w^r$, $g^r$ etc.} (call this function as $\fhat_p(w)$) and (approximately) optimizes it (starting from $w^r$) to get the point $w_p$. Let $d_p=w_p-w^r$. Then $d^r$ is chosen to be any convex combination of $d_p\;\forall p$. Note that, if each $d_p$ is a descent direction then $d^r$ is also a descent direction.

The key is to choose each approximating functional $\fhat_p$ to have {\it gradient consistency} at $w^r$:
\begin{equation}
\fhat_p(w) = \ftilde_p(w) + (g^r - \lambda w^r-\grad L_p(w^r))\cdot(w-w^r)
\label{riskapp}
\end{equation}
Note that $\grad\fhat_p(w^r)=g^r$. This condition gives the necessary tilt to the approximating functions for maintaining consistency with the minimization of the global objective function $f$.

Algorithm 1 gives all the steps of our method. There, $sgd(v^0;\fhat_p,s,pars)$ denotes the output point obtained by applying $s$ epochs of a SGD method to $\fhat_p$, starting from $v^0$, where $pars$ denotes parameters associated with the SGD method, e.g., learning rate. Note that the output point is stochastic since there is randomness present in the iterations, e.g., the order in which examples are presented.

\begin{algorithm2e}
\caption{Distributed method for minimizing $f$ ({\it com:} communication; {\it cmp:} = computation; {\it agg:} aggregation)}
Choose $w^0$; $s$, $pars$, $0\le\theta < \frac{\pi}{2}$\;
\For{$r=0,1 \ldots$}{
1. Compute $g^r$ ({\it com}: $w^r$; {\it cmp:} Two passes over data; {\it agg:} $g^r$); By-product: $\{z_i=w^r\cdot x_i\}$\;
2. Exit if $g^r=0$\;
3. \For{$p=1,\ldots,P$ (in parallel)}{
4.   Set $v^0=w^r$\;
     5. Set $w_p = sgd(v^0;\fhat_p,s,pars)$\;
     6. If $\phase{-g^r,d_p}\ge\theta$, set $d_p=-g^r$; ($\phase{a,b}$ denotes the angle between vectors $a$ and $b$)\;
   }
7. Set $d^r$ as any convex combination of $\{d_p\}$ ({\it agg:} $d_p$)\;
8. Do line search to find $t$ (for each $t$: {\it comm:} $t$; {\it cmp:} $l$ and $\partial l/\partial t$; {\it agg:} $f(w^r+t d^r)$ and its derivative wrt $t$)\;
9. Set $w^{r+1} = w^r+t d^r$\;
}
\end{algorithm2e}

For step 8 we require that the following standard line search conditions are satisfied:
\begin{eqnarray}
\mbox{{\small\bf Armijo:}}\; & f^{r+1} \le f^r + \alpha g^r\cdot(w^{r+1}-w^r) \label{ag} \\
\mbox{{\small\bf Wolfe:}}\;  & g^{r+1}\cdot d^r \ge \beta g^r\cdot d^r \label{wolfe}
\end{eqnarray}
where $0<\alpha<\beta<1$ and $f^r=f(w^r)$.

{\bf Theorem 1.} Algorithm 1 has global linear rate of convergence ({\it glrc}), i.e., $\exists$ $0<\delta<1$ such that $(f(w^{r+1})-f(w^\star)) \le \delta (f(w^r)-f(w^\star))\;\forall r$, where $w^\star=\arg\min_w f(w)$. It follows that algorithm 1 finds a point $w^r$ satisfying $f(w^r)-f(w^\star)\le\epsilon$ in $O(s \log(1/\epsilon))$ time.

Theorem 1 can actually be stated stronger than what is given above. If steps 4-6 of algorithm 1 are replaced by any sub-algorithm that finds a $d_p$ satisfying $\phase{-g^r,d_p}<\theta$ then the glrc result still holds. While convergence follows from standard optimization theory, proving {\it glrc} under such general conditions seems to be a new result. Previously, {\it glrc} seems to have been established only for special cases such as the gradient descent method~\cite{boyd2004}.

If the condition $\phase{-g^r,d_p}\ge\theta$ never gets triggered in step 6, then algorithm 1 can be viewed as a clean parallel SGD method. Step 6 is a ``safe" artifact step that is added to account for the stochasticity of ({\it sgd}). One may ask: {\it how frequently does the condition $\phase{-g^r,d_p}\ge\theta$ happen?}. This can be answered both, from a theoretical as well as a practical angle. Theoretically, we can show that, by making $s$, the number of {\it sgd} epochs in step 5 large, the probability of the condition happening can be made arbitrarily small.

{\bf Theorem 2.} Let $\what_p^\star=\arg\min_w\fhat_p(w)$. Suppose $sgd$ has strong stochastic convergence in the sense that ${\mathbb E}\|w_p-\what_p^\star\|^2 \le K\alpha^s \|w^r-\what_p^\star\|^2\;\forall s$, where $0\le\alpha <1$ and $K\ge 0$. Let $\frac{\pi}{2} > \theta > \cos^{-1}\frac{\lambda}{L}$ where $L$ is a Lipschitz constant for $g$. Then, for any $0<\gamma<1$, $\exists$ $s=O(\log(1/\gamma))$ such that ${\rm Prob}(\phase{-g^r,d^r}\ge\theta)< \gamma$.

Theorems 1 and 2 can be combined to give algorithmic time complexity as $O(\log(1/(\epsilon\gamma)))$.
The proof is based on formalizing the following observations. (1) Since $\fhat_p$ has a `curvature' of at least $\lambda$ and $\grad\fhat_p(w^r)=g^r$, $\what_p^\star-w^r$ makes an angle less than $\cos^{-1}\frac{\lambda}{L}$ with $-g^r$. (2) Strong stochastic convergence of ${\it sgd}$ implies that, as $s$ is made large, $w_p$ comes close to $\what_p^\star$ with high probability. Recent SGD methods~\cite{johnson2013, leroux2012} possess the strong convergence property needed in Theorem 2.

Let us now discuss the practical view of step 6. For {\it sgd}, suppose we use an SGD method such as the ones in~\cite{johnson2013, leroux2012} which move in directions that are ``close" to the negative batch gradient and also have low variance. Then, the points generated by {\it sgd} lie close to the batch negative gradient flow and hence nicely lead to $w_p-w^r$ being a descent direction. This actually happens even when $s$ is small. With large $s$, of course, $w_p$ moves close to $\what_p^\star$ as explained earlier. Thus, the chance of $\fhat_p(w_p)\ge\fhat_p(w^r)$ happening is very low, irrespective of $s$. Note from the definition of $\fhat_p$ that, $w_p-w^r$ is a descent direction of $f$ at $w^r$ if and only if $\fhat_p(w_p)<\fhat_p(w^r)$.

For a practical implementation, we can simply set $\theta=0$, going by standard practice in numerical optimization; thus, directions that lead to descent are accepted. Parameters ({\it pars}) of {\it sgd} can be set as recommended by the individual SGD method used. The number of epochs $s$ can be set based on a communication-computation trade-off; it is wise to choose $s$ so that the cost of communication (of $w^r$, $g^r$ etc.) between nodes is reasonably commensurate with the cost of computation ({\it sgd} epochs) in each node. For step 7, one can use simple averaging. For the line search step we can first calculate $w\cdot x_i$ (note the by-product in step 1) and $d^r\cdot x_i$ for all $i$ in a distributed fashion. Then the calculation of $f(w^r+td^r)$ and its derivative with respect to $t$ is cheap; so one can use any good one dimensional search algorithm to find a point satisfying the Armijo-Wolfe conditions. The parameters in these conditions can be set to: $\alpha=10^{-4}$ and $\beta=0.9$.

\begin{figure}
\centering
\subfigure[25 nodes]{
\includegraphics[width=0.27\linewidth]{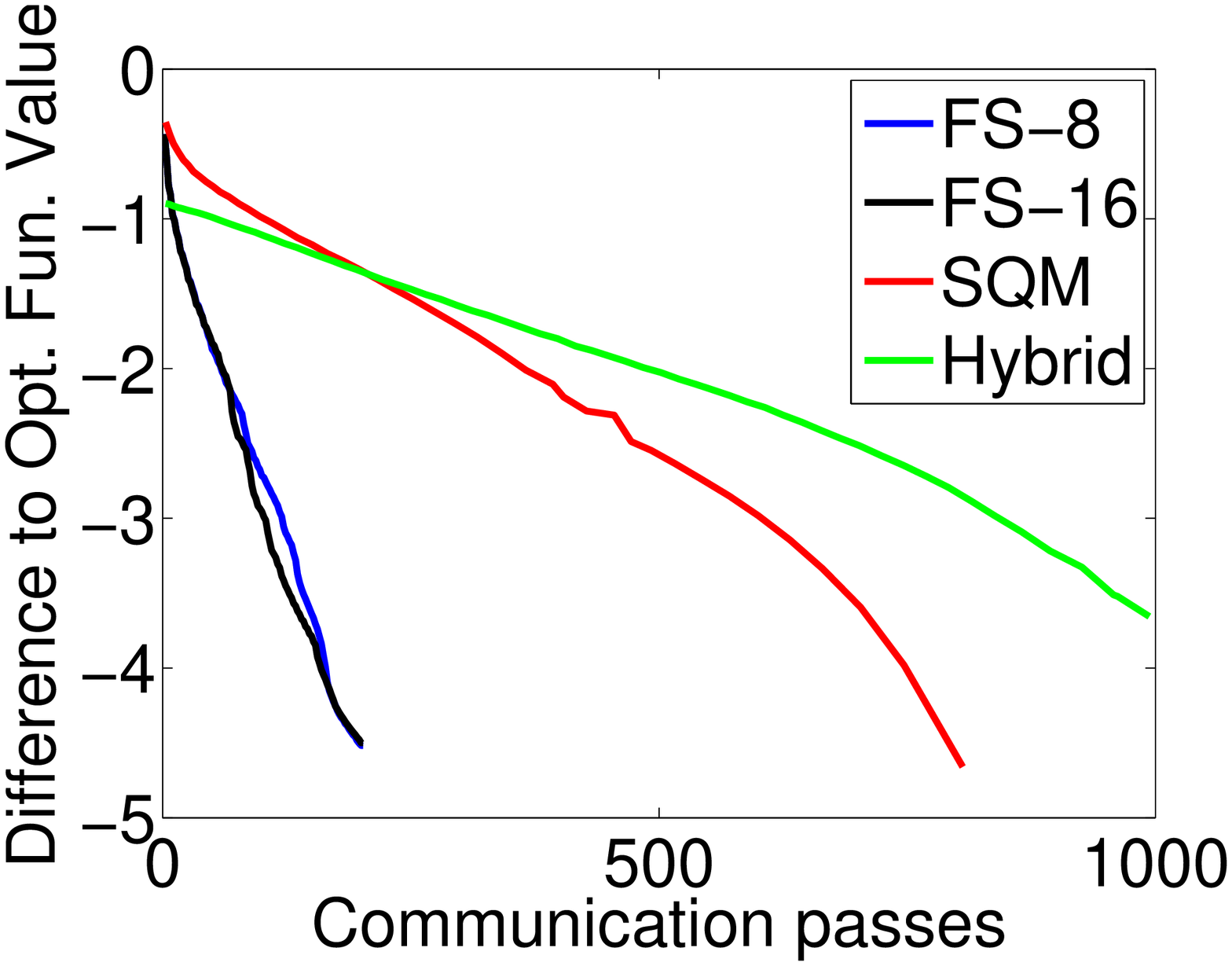}
}
\subfigure[25 nodes]{
\includegraphics[width=0.27\linewidth]{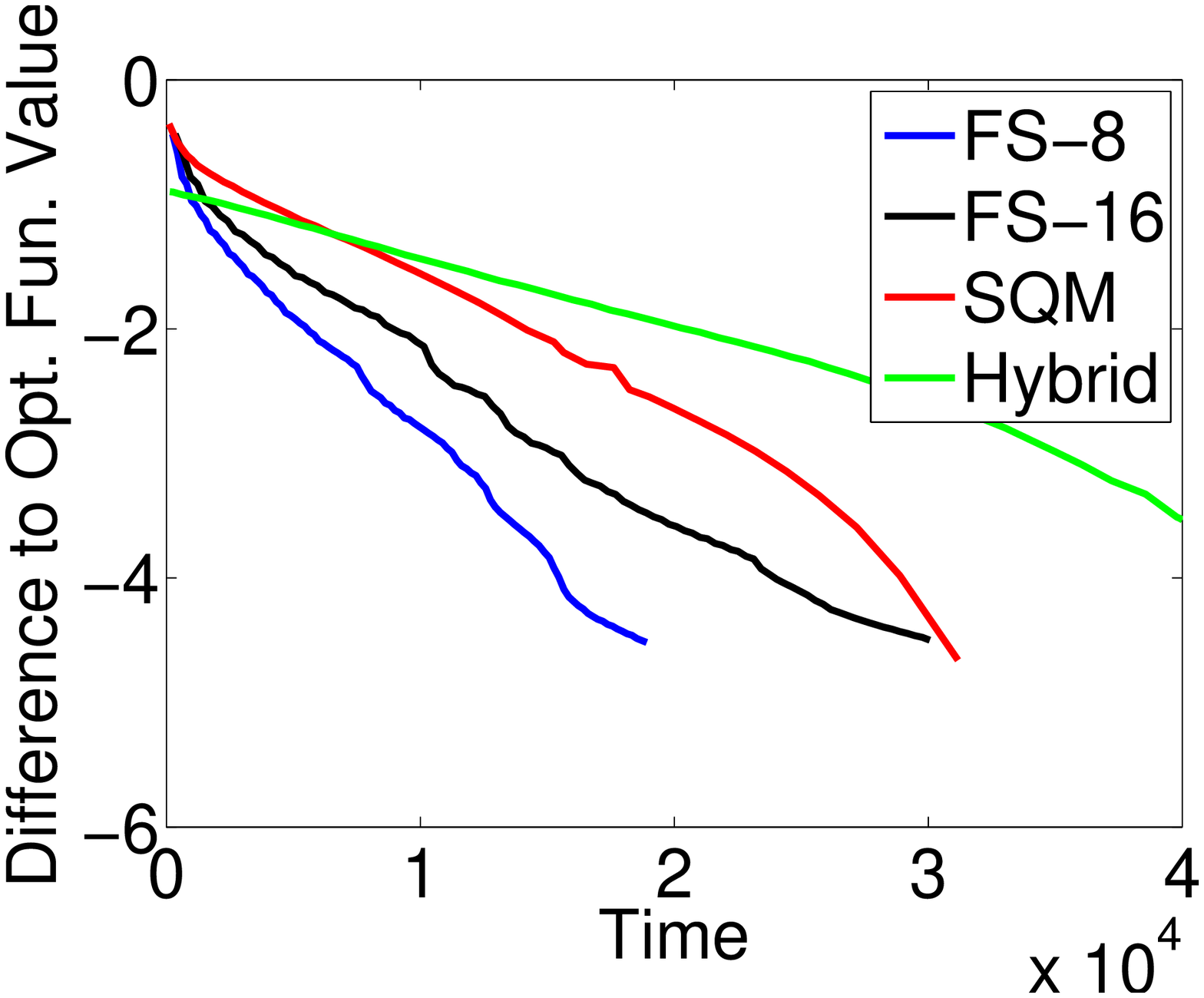}
}
\subfigure[25 nodes]{
\includegraphics[width=0.27\linewidth]{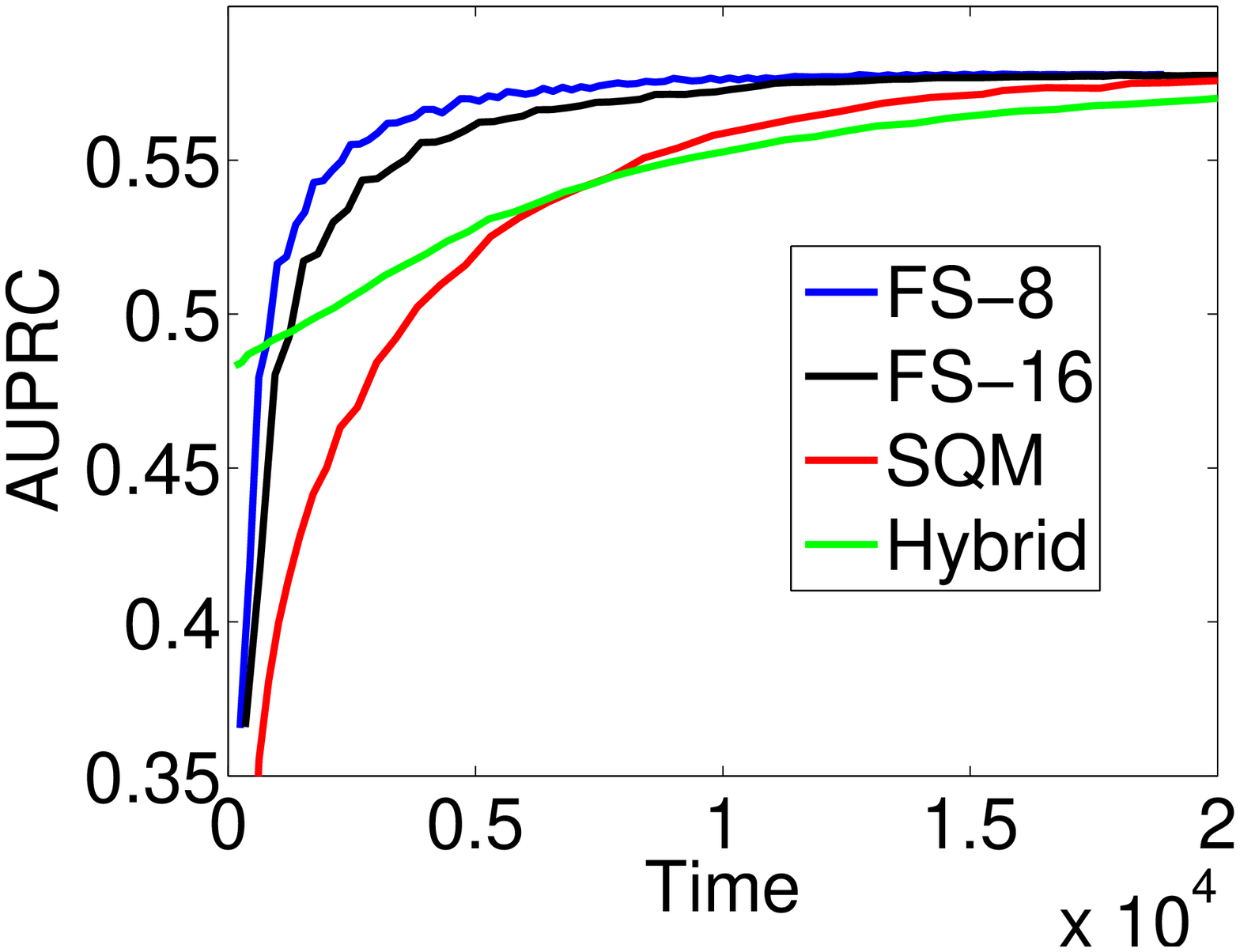}
}
\subfigure[100 nodes]{
\includegraphics[width=0.27\linewidth]{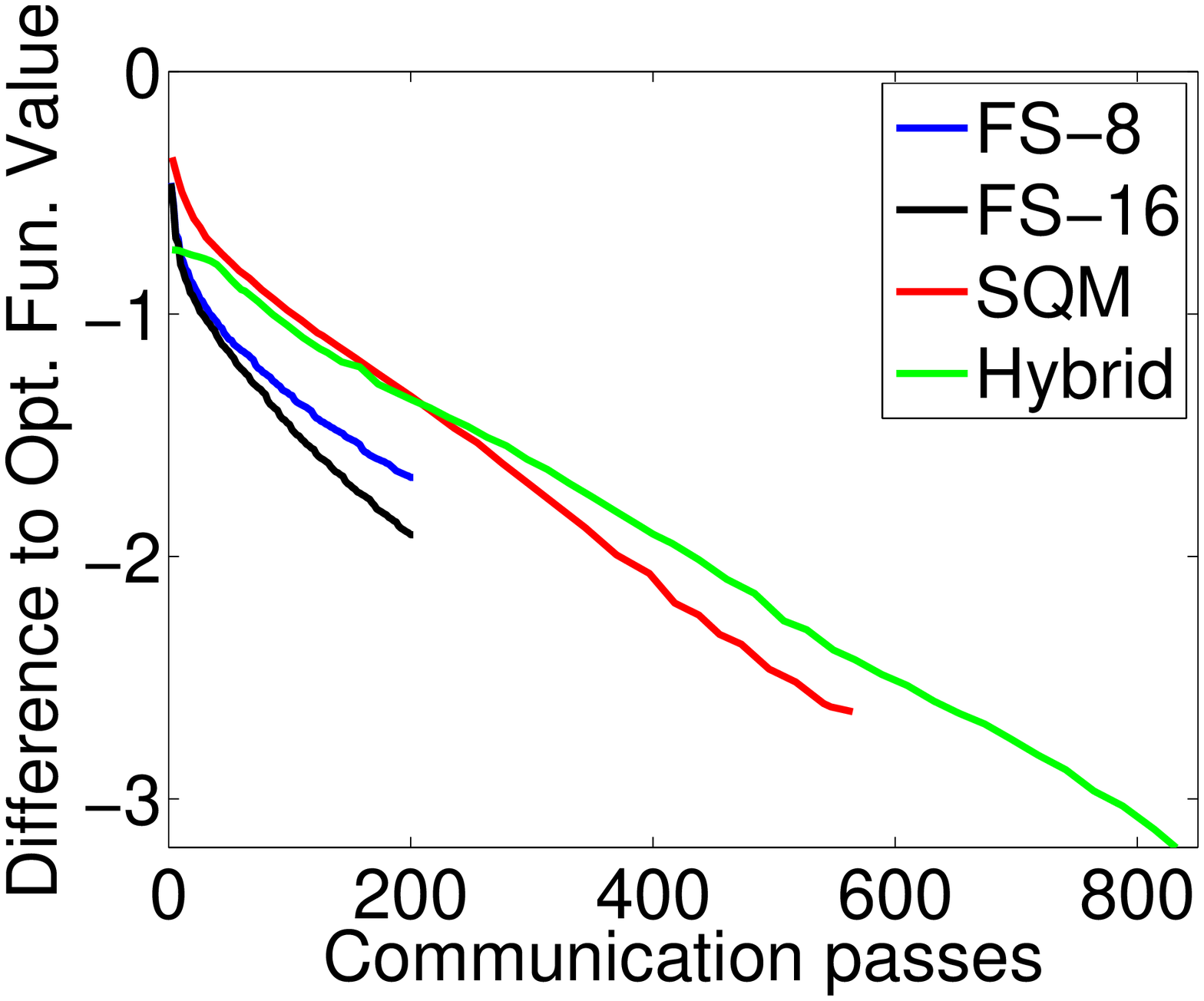}
}
\subfigure[100 nodes]{
\includegraphics[width=0.27\linewidth]{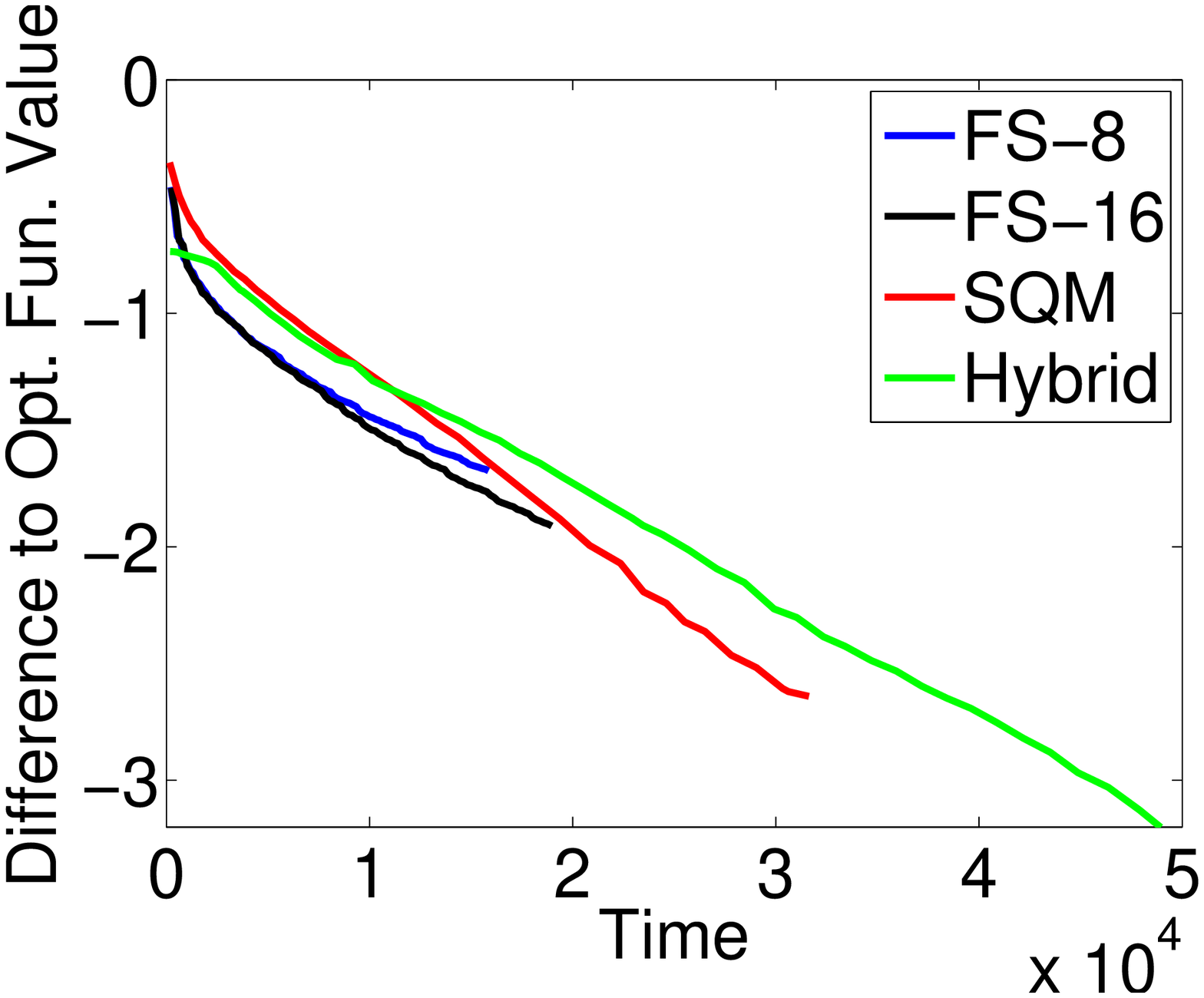}
}
\subfigure[100 nodes]{
\includegraphics[width=0.27\linewidth]{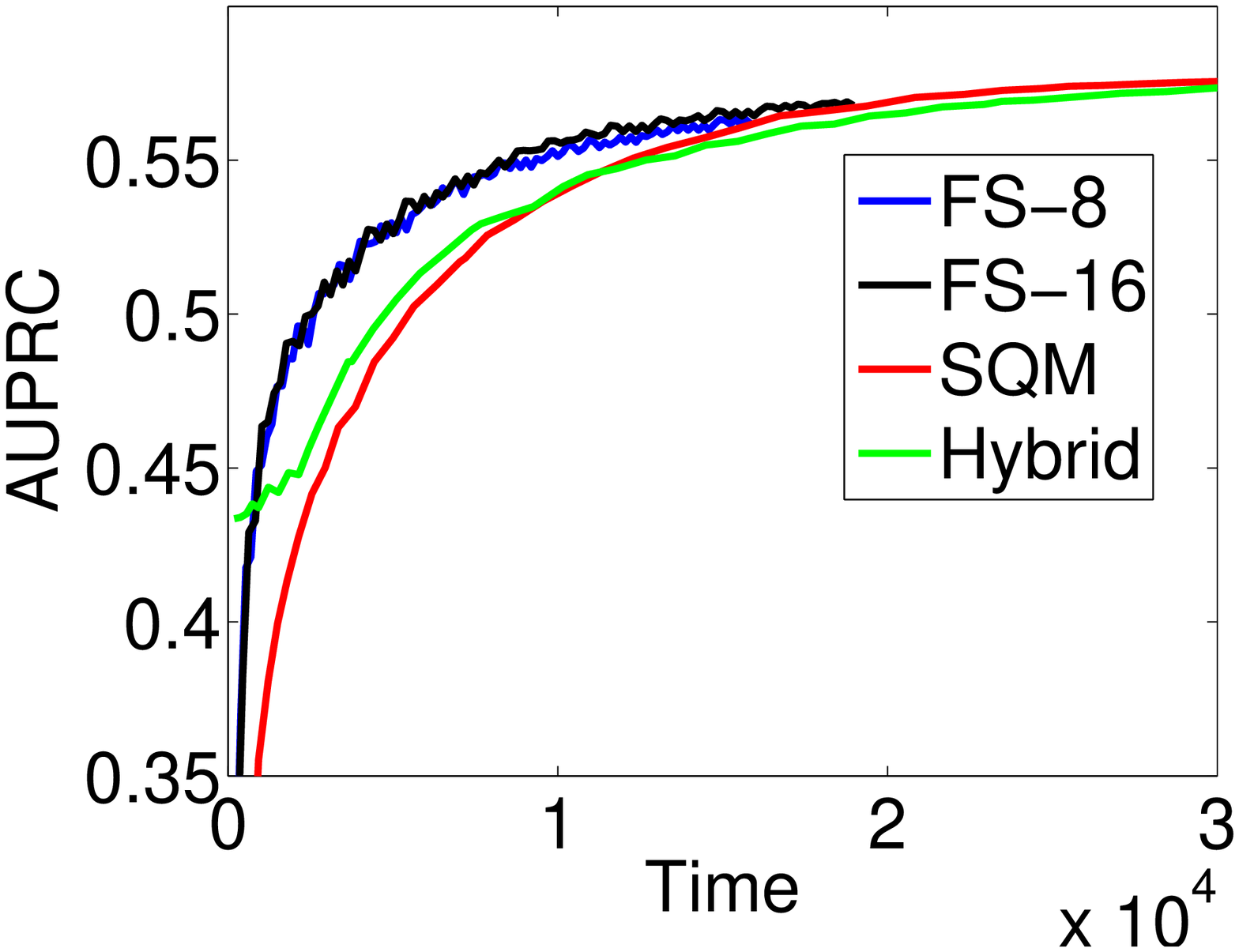}
}
\caption{Comparison of methods. Left and middle: $(f-f^*)/f^*$ ($\log$ scale) versus Number of communication passes and Time (in seconds); Right: AUPRC as a function of Time, for {\it kdd2010} with 25 nodes (top) and 100 nodes (bottom).}
\label{auprc}
\end{figure}

{\bf Experiments.} We use an AllReduce tree running on a Hadoop cluster~\cite{agarwal2011}. We use the Area under Precision-Recall Curve (AUPRC) and $(f-f^*)/f^*$ (in $\log$ scale) as the evaluation criteria. (We obtained $f^*$ by optimizing with very small tolerances to get a very accurate solution.) Experiments are conducted on the {\it kdd2010} dataset in {{\small{\url{http://www.csie.ntu.edu.tw/~cjlin/libsvmtools/datasets/binary.html}}}.
This dataset has 8.41 million examples, 20.21 million examples and 0.3 billion non-zero elements in the data matrix. We use squared hinge loss with $L_2$ regularization. For {\it sgd} we use the SVRG method~\cite{johnson2013}. Let $FS-s$ denote our method with $s$ being the number of {\it sgd} epochs.

We compare our method against {\it SQM} (Statistical Query Model)~\cite{chu2006,agarwal2011}, which is currently one of the most effective distributed methods. {\it SQM} is a batch, gradient-based descent method. The gradient is computed in a distributed way with each node computing the gradient component corresponding to its set of examples, followed by an aggregation of the components via an AllReduce tree. {\it Hybrid} is same as SQM, but uses parameter mixing for initializing. Each node $p$ does one epoch of SGD~\cite{bottou2010} on its examples; then the weights from various nodes are averaged to form a weight vector that is used to initialize {\it SQM}. Our implementations of {\it SQM} and {\it Hybrid} are close to that in~\cite{agarwal2011}; the main difference is that, instead of L-BFGS we use the better-performing TRON~\cite{lin2008} as the core optimizer.

Figure 1 compares our method against {\it SQM} and {\it Hybrid}. First let us look at the variation of the objective function accuracy with respect to the number of communication passes.\footnote{One communication pass corresponds to a vector of size equal to the feature dimension being passed between nodes.} Looking at the plots on the left side of Figure 1 we can see that our method requires far less number of communication passes than {\it SQM} and {\it Hybrid} to achieve the same accuracy in objective function. This difference between the methods also extends to the plots of objective function accuracy versus computing time (middle sub-plots in Figure 1); however, the difference in performance is less pronounced. This is due to the increased computational work ({\it sgd} epochs) done by our method at the nodes; note that {\it SQM} and {\it Hybrid} use the nodes only to compute gradient components.

{\it SQM} and {\it Hybrid} also have the advantage of better convergence when coming close to the optimum since they are directly based on second order modeling of $f$. Our method is good at forming approximate global views of $f$ right from the beginning, thus making good progress in the early iterations. This is also directly reflected in the AUPRC plots (the right side of Figure 1). Our method reaches stable generalization performance much quicker than {\it SQM} and {\it Hybrid}.

When the number of nodes is increased, {\it SQM} and {\it Hybrid} come closer to our method; see this by comparing the sub-plots in the top and bottom of Figure 1. This is due to the fact that, when the number of nodes becomes large, $\fhat_p$ does not approximate $f$ very well, leading to an increased number of major iterations. The value of $s$, the number of SGD epochs plays a key role in determining the rate of linear convergence.

{\bf Conclusion and discussion.} We have given a parallelization of stochastic gradient descent with strong convergence properties and demonstrated its effectiveness.

We can extend our method in several ways.
(a) Suppose $f$ is non-convex, e.g., $f$ from neural networks and deep learning. In the definition of $\fhat_p$ (see (\ref{riskapp}) and (\ref{ftilde})), if $L_p(w)$ is replaced by a convex approximation then convergence of algorithm 1 can be shown; however, complexity results (which require {\it glrc}) are not easy to prove. From a practical point of view, it also makes sense to try non-convex $\fhat_p$, but care is needed to stop the optimization of $\fhat_p$ via {\it sgd} early to make sure that the $d_p$ are descent directions.
(b) For convex $f$, our method can also use other algorithms (e.g., L-BFGS, TRON etc.) as a replacement for {\it sgd} to optimize $\fhat_p$, leading to interesting possibilities. (c) It is also useful to explore automatic ways of switching from our method to {\it SQM} when nearing the optimum.
%the addition of second order terms (e.g., based on L-BFGS approximations) to (\ref{riskapp}) to improve the convergence rate of our method when nearing the optimum.

\bibliography{falda_bib}

\begin{thebibliography}{10}

\bibitem{bottou2010}
L.~Bottou, ``Large-scale machine learning with stochastic gradient descent,''
  in {\em COMPSTAT'2010}, pp.~177--187, 2010.

\bibitem{leroux2012}
N.~Le~Roux, M.~Schmidt, and F.~Bach, ``A stochastic gradient method with an
  exponential convergence rate for strongly convex optimization with finite
  training sets,'' in {\em arXiv}, 2012.

\bibitem{johnson2013}
R.~Johnson and T.~Zhang, ``Accelerating stochastic gradient descent using
  predictive variance reduction,'' {\em NIPS}, 2013.

\bibitem{hsieh2008}
C.~Hsieh, K.~Chang, C.~Lin, S.~Keerthi, and S.~Sundararajan, ``A dual
  coordinate descent method for large-scale linear svm,'' in {\em ICML},
  pp.~408--415, 2008.

\bibitem{zinkevich2010}
M.~Zinkevich, M.~Weimer, A.~Smola, and L.~Li, ``Parallelized stochastic
  gradient descent,'' in {\em NIPS}, pp.~2595--2603, 2010.

\bibitem{mann2009}
G.~Mann, R.~McDonald, M.~Mohri, N.~Silberman, and D.~Walker, ``Efficient
  large-scale distributed training of conditional maximum entropy models,'' in
  {\em NIPS}, pp.~1231--1239, 2009.

\bibitem{hall2010}
K.~Hall, S.~Gilpin, and G.~Mann, ``Mapreduce/bigtable for distributed
  optimization,'' in {\em NIPS Workshop on Leaning on Cores, Clusters, and
  Clouds}, 2010.

\bibitem{agarwal2011}
A.~Agarwal, O.~Chapelle, M.~Dudik, and J.~Langford, ``A reliable effective
  terascale linear learning system,'' in {\em arXiv}, 2011.

\bibitem{boyd2004}
S.~Boyd and L.~Vandenberghe, {\em Convex optimization}.
\newblock Cambridge, UK: Cambridge University Press, 2004.

\bibitem{chu2006}
C.~Chu, S.~Kim, Y.~Lin, Y.~Yu, G.~Bradski, A.~Ng, and K.~Olukotun, ``Map-reduce
  for machine learning on multicore,'' {\em NIPS}, pp.~281--288, 2006.

\bibitem{lin2008}
C.~Lin, R.~Weng, and S.~Keerthi, ``Trust region newton method for large-scale
  logistic regression,'' {\em JMLR}, pp.~627--650, 2008.

\end{thebibliography}
\bibliographystyle{ieeetr}

\end{document}